\documentclass[runningheads]{llncs}

\usepackage[T1]{fontenc}
\usepackage{graphicx}
\graphicspath{{figures/}}

\usepackage{amsmath,amssymb}
\usepackage{booktabs}
\usepackage{xcolor}
\usepackage{hyperref}
\usepackage{enumitem}
\usepackage{mdframed}
\usepackage{placeins}
\usepackage{silence}
\usepackage{subcaption}

\hypersetup{colorlinks=true, linkcolor=blue, citecolor=blue, urlcolor=blue}

\newcommand{\selfcitehodson}{\cite{hodson2026sl}}

\newcommand{\attn}{attentive agent}
\newcommand{\Attn}{Attentive agent}
\newcommand{\base}{uniform agent}
\newcommand{\Base}{Uniform agent}
\newcommand{\rev}{anti-aligned control}
\newcommand{\Rev}{Anti-aligned control}
\newcommand{\inferonly}{inference-only ablation}

\newcommand{\authorcid}[1]{}

\begin{document}

\title{Interoceptive Attention as Dynamic Homeostatic Prioritization in a Foraging Agent}
\titlerunning{Interoceptive Attention as Homeostatic Prioritization}

\author{
St John Grimbly\inst{1,2}\authorcid{0000-0002-3003-1815}\thanks{Corresponding author: uct@stjohngrimbly.com} \and
Nicolas Kuske\inst{1,2}\authorcid{0000-0002-1054-6811} \and
Evert A. Boonstra\inst{2} \and
Bruce A. Bassett\inst{1,4} \and
Charel van Hoof\inst{6} \and
Rowan Hodson\inst{3} \and
Benjamin Rosman\inst{4} \and
Ryan Smith\inst{3} \and
Mark Solms\inst{2}\authorcid{0000-0003-4828-3882}\thanks{Co-senior authors.} \and
Jonathan P. Shock\inst{1,2,5}\authorcid{0000-0003-3757-0376}\protect\footnotemark[\value{footnote}]
}
\authorrunning{Grimbly et al.}
\institute{
Dept.\ of Mathematics \& Applied Mathematics, Univ.\ of Cape Town, South Africa \and
Neuroscience Institute, Univ.\ of Cape Town, South Africa \and
Laureate Institute for Brain Research, Tulsa, OK, USA \and
MIND Institute and CSAM, University of the Witwatersrand, South Africa \and
INRS, Montreal, Canada \and
Delft University of Technology, Dept.\ of Cognitive Robotics
}

\maketitle

\begin{abstract}
\begin{sloppypar}
Biological systems must regulate competing needs under limited
perceptual bandwidth, where sharpening one estimate costs the
capacity to sharpen the others. Any fixed-budget system therefore
has to decide where to allocate its perceptual precision. We study
this in a foraging agent that must keep several bodily needs
satisfied to survive, modelled with active inference. At each step
it reads its own body-state beliefs, identifies the most-needed
channel, and reallocates a fixed budget of interoceptive
\emph{precision} toward it, so that the same precision-shaped
likelihood feeds both belief update and planning. In
\textsc{AffectWorld}, a four-channel foraging gridworld, this
selective allocation more than doubles learning-phase survival at
matched budget against a uniform-precision agent ($0.414$ vs
$0.199$ across 11 layouts, $n{=}32$ seeds each, paired
cluster-bootstrap $p \leq 10^{-4}$). Two further results sharpen
the mechanism. The benefit runs through planning as well as
perception, since denying the shaped likelihood to the planner
alone removes about half of it. It is also need-aligned, since
aiming precision at the least-needed channel does worse than
spreading it evenly. The attended channel additionally learns its
own dynamics about twice as fast, and stays ahead even at matched
observation count, a behavioural trace of the same precision
routing, visible in learning speed, not survival.
\end{sloppypar}

\end{abstract}

\section{Introduction}
\label{sec:intro}

A foraging animal must keep several bodily needs satisfied at
once. It has to eat, drink, and breathe, and letting any one of
these lapse too far is fatal. Faced with many needs and one body,
the animal must continually decide what to do next, and
behavioural ecology and the animat tradition have long studied
this as a problem of \emph{action selection} among competing
needs
\cite{mcnamara1986common,maes1991bottom,canamero2003designing,lewis2016hedonic}.
A prior question has drawn less attention. Before choosing an
action, the animal must decide which of its internal signals to
trust, because it cannot monitor every need equally well at every
moment, and attending closely to one signal leaves less capacity
for the rest. This paper is about that perceptual choice: how an
agent should allocate limited sensory precision across competing
interoceptive channels, and what such allocation buys it.

The quantity being allocated is \emph{precision}: the weight an
agent gives a signal when deciding how much to trust it. In
predictive coding and active inference, precision is the confidence
assigned to a prediction error, and attention is modelled as
control over it
\cite{bastos2012canonical,feldman2010attention,parr2017uncertainty}.
Cognitive science studies the same trade-off as bounded or
resource-rational attention \cite{lieder2020resource}, a limited
processing budget spread across inputs. Because that budget is
fixed, trusting one signal more means trusting the others less. In
our model this allocation plays two roles, and we treat them
separately: it changes how clearly the agent senses whichever need
is most urgent, and it changes which actions the agent then judges
worthwhile.

\begin{figure}[t]
\centering
\begin{subfigure}[t]{0.55\textwidth}
\centering
\includegraphics[width=\linewidth]{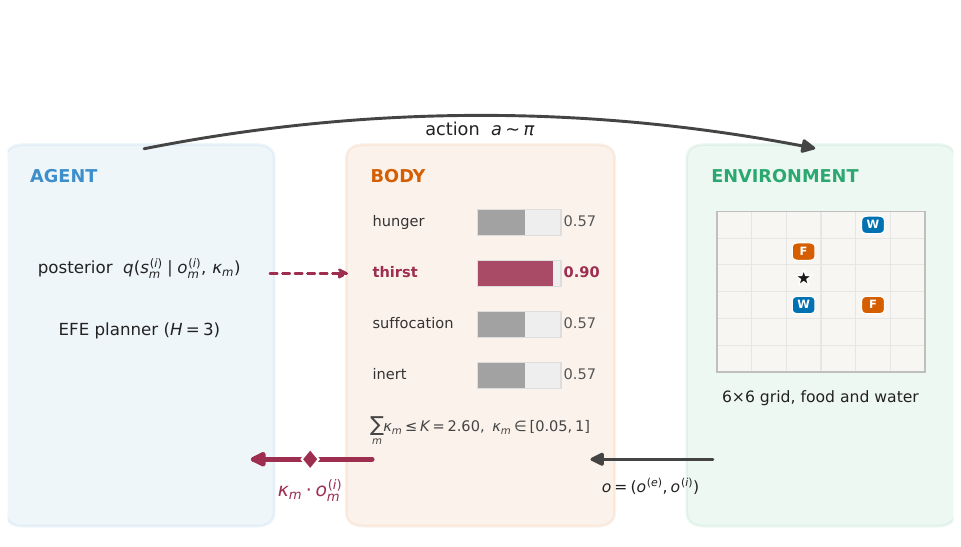}
\caption{}
\label{fig:schematic}
\end{subfigure}\hfill
\begin{subfigure}[t]{0.43\textwidth}
\centering
\includegraphics[width=\linewidth]{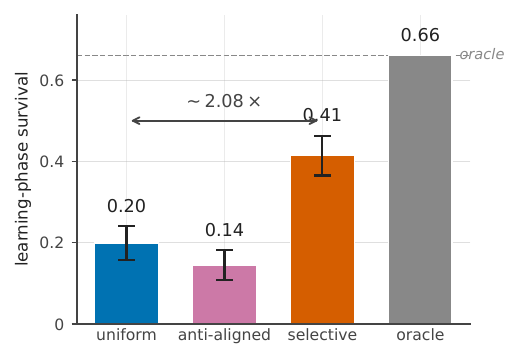}
\caption{}
\label{fig:teaser-results}
\end{subfigure}
\caption{\textbf{Architectural intervention and headline result.}
\textbf{(a)}~One shaped likelihood field $A^{(m)}$ with two
downstream consumers, the belief update and the EFE planner,
under the budget $\sum_m \kappa_m \leq K{=}2.60$. Suffocation is
tied to water.
\textbf{(b)}~Across 11 layouts ($n=32$ seeds), selective
allocation more than doubles (${\sim}2.08{\times}$) uniform survival. The anti-aligned
control reverses the gain. Oracle ceiling: a planner given the true
environment model.}
\label{fig:teaser}
\end{figure}

This could be pursued in several modelling frameworks.
Reinforcement learning, for instance, can be given interoceptive
state and a homeostatic reward. We use active inference for two
reasons. Precision is already a first-class quantity in it, so the
mechanism we care about is native rather than added on. And it has
become a standard setting for biologically motivated, Bayesian
models of decision making, including homeostatic and interoceptive
control
\cite{pezzulo2015active,seth2016active,tschantz2022simulating}. In
active inference \cite{friston2017active,parr2022active} the agent
chooses actions to make its observations match its predictions,
under a single objective defined over both its beliefs and its
plans. In our model the precision-shaped likelihood enters both the
belief update and the evaluation of plans (Fig.~\ref{fig:teaser}a),
so one precision choice acts on perception and decision at the same
time. Whether that dual action helps, and where, is a question our
propagation analysis addresses.

Our contribution sits between three lines of work. Homeostatic
reinforcement learning integrates bodily state through a
hand-designed reward \cite{keramati2014,yoshida2024homeostatic},
but routes behaviour through reward rather than precision.
Active-inference models of interoception use precision to regulate
homeostatic control \cite{pezzulo2015active,seth2016active}, but
not as a variable the agent dynamically reallocates across
competing interoceptive channels. Active-inference treatments of
attention as precision have mostly addressed a single
exteroceptive target, such as visual search
\cite{parr2017uncertainty,mirza2019introducing}, or cast the
control of precision as an action within a hierarchical model
\cite{whyte2022access}, rather than spreading a fixed budget across
several interoceptive needs. We combine
these: a fixed interoceptive precision budget, routed step by step
by the agent's own belief about which need is most urgent, and used
at both the perception and planning stages.

\begin{sloppypar}
We ask a direct question: does pointing a limited precision budget
at whichever need is currently most urgent help an agent survive
and learn, compared with spreading precision evenly? We test this
in \textsc{AffectWorld}, a foraging gridworld with several
competing needs, and the answer is yes. Selective allocation
roughly doubles learning-phase survival against a uniform-precision
agent at the same budget (Fig.~\ref{fig:teaser}b). The
\emph{direction} of allocation is what matters. An agent that
instead sharpens its least-needed channel does worse than uniform,
so the gain comes from tracking need, not from unevenness on its
own. Two further results locate the effect. The precision-shaped
signal helps at both the perception and planning stages, and
denying it to the planner alone costs about half the benefit. The
attended channel also learns its own dynamics about twice as fast,
staying ahead even at matched observation count, a behavioural
difference between the two agents that gives another view of the
mechanism at work.
\end{sloppypar}

\section{Mechanism and Setting}
\label{sec:setting}

\begin{figure}[t]
\centering
\includegraphics[width=0.85\textwidth]{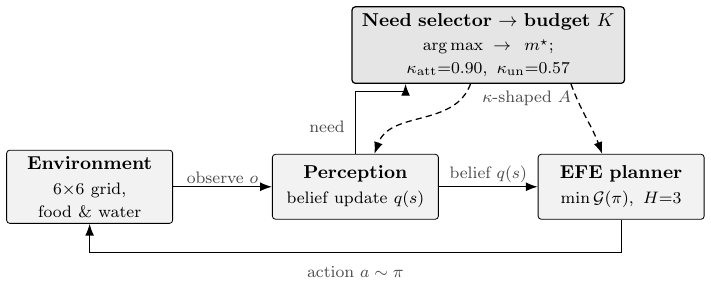}
\caption{\textbf{The $\kappa$-attention loop.} The agent observes its
position, the local resource, and four interoceptive channels, updates
its body-state belief, and attends the most-needed one. That choice
sets a fixed precision budget $K$ whose shaped likelihood $A^{(m)}$
enters both the belief update and the EFE planner (dashed) before the
planner acts and closes the loop.}
\label{fig:loop}
\end{figure}

\subsection{Active Inference under Fixed Interoceptive Precision}
\label{sec:actinf}

The agent is an active-inference agent: it maintains beliefs about
its bodily and world state and acts to make its observations match
its predictions (Fig.~\ref{fig:loop}). Formally, it maintains a factorised partially
observable Markov decision process (POMDP), a natural model class
for an agent whose interoceptive observations are noisy reports of
underlying physiological states, and selects policies by minimising
expected free energy \cite{friston2017active,parr2022active}. In its
joint form
\[
\mathcal{G}(\pi) \;=\; \mathbb{E}_{q(s, o\mid\pi)}\!\bigl[\log q(s\mid\pi) - \log P(s, o\mid C)\bigr],
\]
\begin{sloppypar}
this decomposes into a \emph{risk} term
$\mathbb{E}_{q(o\mid\pi)}\!\bigl[D_{\mathrm{KL}}\!\bigl(q(o\mid\pi)\,\|\,P(o\mid C)\bigr)\bigr]$
(divergence between predicted and preferred observations) and an \emph{ambiguity}
term $\mathbb{E}_{q(s\mid\pi)}\!\bigl[\mathcal{H}[P(o\mid s)]\bigr]$
(expected entropy of the observation likelihood). In words, a
policy scores well when it is expected to move the body toward its
preferred replete states (low risk) while keeping observations
informative about body state (low ambiguity). The preference
distribution $P(o\mid C)$ is a soft prior concentrated near the
full body-state level ($s{=}s_{\max}$) on each channel. Policies are sampled from
$P(\pi){\propto}\exp(-\gamma\mathcal{G}(\pi))$ with $\gamma{=}16$
fixed throughout. Precision $\kappa$ enters both EFE
terms through the likelihood $A^{(m)}$ defined in
Sec.~\ref{sec:kappa-mechanism}. The agent's generative model has two parts: a fixed
body-transition prior $B$ (matching biology's access to body
dynamics through proprioception and prior experience) and a
learned observation likelihood $A$ mapping body and world states
to observations. $A$ is updated online via Dirichlet pseudo-counts
on each observation (Sec.~\ref{sec:relevant}). Its Dirichlet prior has a concentration $\alpha_0$, and a larger $\alpha_0$ makes the model more rigid, so observations move the body-state posterior less. The agent therefore
learns what each interoceptive reading reports about its true body
state and what each exteroceptive reading reports about the grid.
Actions are the four cardinal moves plus a no-op (stay-in-place).
Resource consumption on a tile is automatic.
\end{sloppypar}

The body produces a noisy categorical observation
$o^{(i)}_m \in \{0,\ldots,5\}$ on each of $M{=}4$ interoceptive
channels $m \in \{1, \ldots, 4\}$ (three active needs and one inert
control) at observation precision
$\kappa_m$. Concretely, $\kappa_m$ is the probability that the
per-step observation equals the underlying body-state level on
channel $m$ — the diagonal of the channel's likelihood matrix is
$\kappa_m$, with each ``wrong'' read drawn uniformly over the five
other levels:
\[
A^{(m)}_{o,\,s} \;=\;
\begin{cases}
\kappa_m & \text{if } o = s, \\[2pt]
(1 - \kappa_m)/5 & \text{otherwise,}
\end{cases}
\qquad o,\,s \in \{0, \ldots, 5\},
\]
so each column of $A^{(m)}$ is a valid categorical distribution
and $\kappa_m \in [0,1]$ is the per-step probability that the
observation reports the true level on channel $m$. Total
interoceptive precision is constrained to a fixed soft
budget,\footnote{We set $K = 2.60$ so that the attended channel is
informative while the others are not blind.
Sec.~\ref{sec:selectivity} sweeps $K$ over $\{1.5,\ldots,4.0\}$.}
\[
\sum_{m=1}^{4} \kappa_m \,\leq\, K, \qquad
\kappa_m \in [\kappa^{\text{floor}},\,1], \qquad
K = 2.60, \quad \kappa^{\text{floor}} = 0.05,
\]
where the floor prevents any channel from being completely
silenced and the upper bound $\kappa_m \leq 1$ is required for
$A^{(m)}_{o,s}$ to remain a valid probability. The allocation
procedure clips per channel,
$\kappa_m \leftarrow \mathrm{clip}(\tilde\kappa_m, \kappa^{\text{floor}}, 1)$,
and any residual budget is absorbed into the constraint
$\sum_m\kappa_m\le K$. Uniform allocation gives
$\kappa_m = K/4 = 0.65$ on every channel — each channel reports
correctly $65\%$ of the time. Selective allocation gives
$\kappa_{\text{att}} = 0.90$ ($90\%$ correct on the attended
channel) and $\kappa_{\text{un}} = (K - \kappa_{\text{att}})/3
\approx 0.567$ ($56.7\%$ correct on each of the other three),
preserving the sum.

\subsection{The $\kappa$-Attention Mechanism}
\label{sec:kappa-mechanism}

We use \emph{attention} here in a narrow, operational sense. It
means the agent's ongoing allocation of a fixed interoceptive
precision budget across channels, set by its own belief about which
need is most urgent, and it is a computational abstraction rather
than a model of a specific neural system. In the regime we study
the body-state posterior stays well-calibrated to true need, so the
allocation tracks genuine urgency.

\begin{sloppypar}
The selector reads only the agent's own posterior beliefs
$q(s^{(i)}_m)$, never ground-truth body counters. The body
observation model is itself learned online, so body beliefs
remain uncertain throughout. At each step the agent picks an
attended channel $m^\star_t$ from its own posterior over body
states,
$m^\star_t = \arg\max_m \mathbb{E}_{q(s^{(i)}_m)}[\mathrm{need}_m]$,
and reallocates the precision budget to give that channel the
higher precision $\kappa_{\text{att}}$, so the agent sharpens its
interoceptive sense on whichever need it currently believes is most
pressing. The scalar
$\mathrm{need}_m(s) = (s_{\max} - s)/s_{\max}$ for the three
active channels (full body-state level $\to 0$, empty $\to 1$),
and returns a constant $0$ on the inert control channel. Ties
are broken by lowest channel index. The default rule
(need-aligned, picking the most-needed channel) is used in all
main results. Alternative criteria (action-aware, explorative,
hysteresis, anti-aligned) appear as ablations in
Sec.~\ref{sec:direction}, along with a ground-truth selector
(reading true body state) as a directional ceiling on a
single-layout pilot ($n=3$).
\end{sloppypar}

The $\kappa$ allocation could act at three sites in the agent:
(i)~\emph{belief update} ($A^{(m)}$ enters per-step state
inference and breaks posterior symmetry in a need-correlated
way), (ii)~\emph{policy evaluation} (the same $A^{(m)}$
enters EFE and sharpens the predicted-vs-preferred KL on that
channel), and (iii)~\emph{likelihood learning} (cleaner
observations on the attended channel produce more concentrated
Dirichlet updates of the $A^{(m)}$ pseudo-counts). We treat (iii) as an observed
consequence of (i), since the update rule itself is unchanged across
all agents, and test the contribution of (i) versus (ii) in
Sec.~\ref{sec:mechanism}.

\subsection{Environment and Agents}
\label{sec:env-agents}

\begin{figure}[!tb]
\centering
\includegraphics[width=0.62\textwidth]{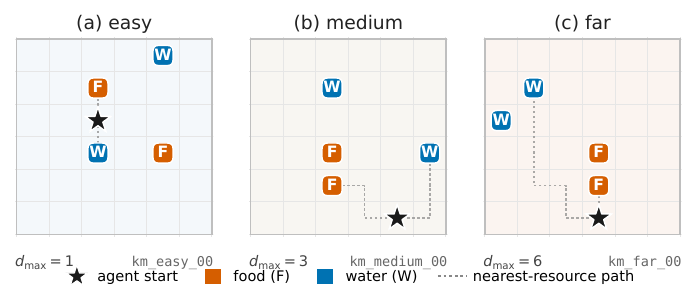}
\caption{Representative easy, medium, and far tier layouts of
\textsc{AffectWorld}, with start-to-resource distances annotated.
The channels-to-resource mapping and budget split are summarised
in the Sec.~\ref{sec:env-agents} prose.}
\label{fig:env}
\end{figure}

\paragraph{Environment.} \textsc{AffectWorld} is a $6{\times}6$
gridworld with two food and two water tiles per layout. Agents
receive two exteroceptive observations (position and resource)
and three interoceptive observations, one per active body
channel (hunger, thirst, suffocation). Episodes run up to 60 steps, terminating early on death. We use 12 distinct environment
layouts across easy, medium and far tiers (Fig.~\ref{fig:env}).
The headline results use 11 (the excluded \texttt{L01} is a
mechanistic stress-test).

\paragraph{Body Channels.} Hunger, thirst and suffocation are
three distinct needs, each starting the trial in the \emph{full}
state. Hunger and thirst are relieved by the food and water tiles
respectively, decay one unit/step, and are lethal at zero.
Suffocation depletes on water tiles, recovers elsewhere, and is
non-lethal: zero suffocation lowers the value the agent assigns
to staying on water, but does not terminate the trial. This
gives a need the agent should prioritise during water-seeking
without introducing a second mortality route alongside hunger
and thirst. The fourth channel is an inert control: its underlying
signal is constant, so it carries no need information and the
need-aligned selector has no reason to attend it. It anchors the
primary \attn{}-vs-\base{} comparison: a channel that reports
nothing about any need cannot by itself create the survival gap
between those two agents.

\paragraph{Agents.} The three core agents share planner, priors,
Dirichlet hyperparameters, body model and exteroceptive observation
structure, and differ only in $\kappa$ allocation. The planner
minimises EFE over the full policy space at planning horizon
$H{=}3$ throughout. The \base{}
holds $\kappa = 0.65$ uniformly across all four channels. This fixes
the total precision budget at $K = M \times 0.65 = 2.60$ for
$M{=}4$ channels, and the attentive and reversed agents redistribute
this same total, so all three agents are compared at an identical
budget. The \attn{} runs dynamic $\kappa$
with $\kappa_{\text{att}} = 0.90$ on whichever channel the
body-state belief currently flags as most-needed and
$\kappa_{\text{un}} = 0.567$ on the other three, preserving the
fixed budget. The \rev{} uses the same $\kappa$ split
and same budget but selects the \emph{least}-needed channel, a
direction-flipped control.

\paragraph{Metrics and Grouping.} A \emph{trial} is one episode
(up to 60 steps), scored $1$ if the agent survives to the $60$-step
horizon and $0$ if it dies first. A \emph{run} is one seed's
sequence of $100$ trials on a fixed layout, over which the
likelihood $A$ is learned online. Each (layout, agent) cell uses
$n{=}32$ seeds unless a caption states otherwise.
\emph{Learning-phase survival} is the mean trial outcome over a
run's $100$ trials, pooled across seeds and layouts.
\emph{Plateau survival} restricts that mean to trials $10$--$20$,
once both agents reach steady state. The headline result pools the
11 layouts across easy, medium and far tiers. The \emph{easy-tier}
grouping is the five remaining easy-tier layouts.

\section{Results}
\label{sec:results}

\subsection{Does Selective Precision Beat Uniform?}
\label{sec:headline}

\begin{table}[t]
\centering
\caption{Learning-phase survival across 11 layouts (\texttt{L01}
excluded), $n{=}32$ seeds per cell. Cluster-bootstrap $95\,\%$ CIs
and raw paired-bootstrap $p$-values (Holm-corrected in
Sec.~\ref{sec:limitations}).\protect\footnotemark}
\label{tab:headline}
\begin{tabular}{lccc}
\toprule
Agent & Survival rate & 95\% CI & $p$ vs \base{} \\
\midrule
\Base{}    & $0.199$          & $[0.158, 0.240]$ & --- \\
\Attn{}    & $\mathbf{0.414}$ & $[0.365, 0.463]$ & $\mathbf{\leq 10^{-4}}$ \\
\Rev{}     & $0.144$          & $[0.108, 0.181]$ & $\mathbf{0.004}$ \\
\bottomrule
\end{tabular}
\end{table}
\footnotetext{Code, layout banks, and analysis pipelines for this paper are
available at
\url{https://github.com/sgrimbly/attention-aif-sab2026-snapshot},
where the supplementary appendices provide fuller experiments,
results, and interpretations.}

Replacing uniform $\kappa$ with a body-state-driven selector at the
same budget more than doubles learning-phase survival
(${\sim}2.08\times$, paired-bootstrap $p \leq 10^{-4}$,
Table~\ref{tab:headline}). The \rev{}, which uses the same budget
and allocation magnitudes but flips the selector direction,
performs significantly \emph{worse} than baseline. The task is hard in this setting. Even the oracle planner,
given the true environment dynamics, reaches only ${\sim}0.66$
survival, so we treat it as a performance ceiling rather than a
competing baseline.

\subsection{Does the Direction of Allocation Matter?}
\label{sec:direction}

\begin{figure}[!tb]
\centering
\includegraphics[width=0.55\textwidth]{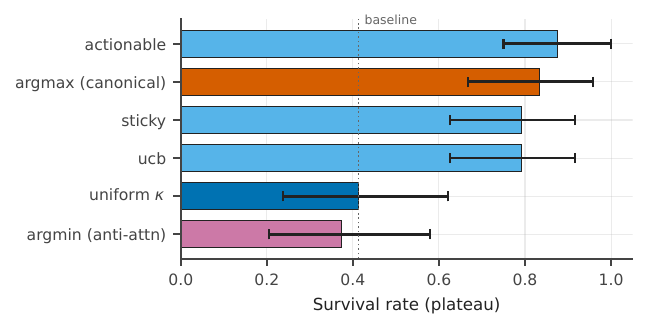}
\caption{Every need-aligned selector beats the \base{} and only
the anti-aligned rule fails, so the advantage comes from the
direction of allocation, not from unevenness alone
($3$ layouts $\times$ $8$ seeds $\times$ $30$ trials, plateau
survival, bootstrap $95\,\%$ CIs).}
\label{fig:direction}
\end{figure}

We probe selector choice and allocation magnitude jointly
(Fig.~\ref{fig:direction}). Four need-aligned criteria plus a
ground-truth selector substantially beat the \base{}. Only the
direction-flipped anti-aligned rule fails. The mechanism is
robust to the specific criterion as long as the selector is
direction-aligned.

Direction is what does the work, not allocation magnitude alone.
Sweeping $\kappa_{\text{att}}$ jointly with the selector at
fixed $K$, the need-aligned rule climbs
monotonically with asymmetry while the anti-aligned rule stays
flat near $0.36$. The two converge to
baseline at uniform allocation ($\kappa_{\text{att}}=0.65$) and
diverge thereafter, reaching a $+44$\,pp gap at the default
$\kappa_{\text{att}}=0.90$ and $+46$\,pp at
$\kappa_{\text{att}}=0.99$.

\subsection{Where Does the Precision Signal Act?}
\label{sec:mechanism}

A single precision-shaped likelihood matrix $A^{(m)}$ is read
by two parts of the agent, the per-step belief update and the
EFE planner. To test how much of the gain rides on the planner
also seeing the shaped likelihood, we run an \inferonly{} variant
that gives the planner an \emph{unshaped} likelihood, with
$A^{(m)}$ rebuilt using the uniform allocation $\kappa_m = K/M$ on
every channel so that likelihood precision is decoupled from
posterior urgency at the planning stage. A drop here is the gain
that the planner's shaped likelihood was supplying.

\begin{sloppypar}
At the default prior
concentration ($\alpha_0{=}0.1$) the \inferonly{} loses
$20$\,pp relative to the full \attn{}. The inversion begins
at $\alpha_0{\geq}1$, and by $\alpha_0{=}10$ the loss reaches
$88$\,pp, with the \inferonly{} collapsing to the level of
the direction-flipped \rev{}. When the body-state posterior is
dominated by a rigid prior, only the planner's use of the shaped
likelihood can push policies away from the prior's preferred
actions. The inference site alone cannot. This refines the
active-inference picture of attention as policy-precision
modulation
\cite{parr2017uncertainty,feldman2010attention,seth2016active}.
State-dependent precision modulation of the observation
likelihood, formalised for exteroceptive visual search
\cite{mirza2019introducing}, here acts on multi-channel
interoceptive precision under a fixed budget and propagates
from inference into the EFE planner.
\end{sloppypar}

The mirror ablation (\emph{planning-only}, with inference-stage
shaping disabled and the planner keeping the shaped $A$) on the
same five easy-tier layouts reaches pooled survival $0.865$, not
significantly above the full model's easy-tier $0.696$ given the
smaller planning-only sample (overlapping CIs). At loose priors the planning-stage
pathway therefore carries the gain on its own, and the
inference-stage shaping is at most neutral here.
Combined with the \inferonly{} collapse at rigid priors, this
localises the dominant pathway in the planner, with inference-stage
shaping becoming necessary only once the prior is strong enough to
override state inference.

\subsection{Is the Advantage Robust to Budget and Prior Rigidity?}
\label{sec:selectivity}

The benefit holds across the three parameters that most plausibly
drive it: prior rigidity $\alpha_0$, attended-channel precision
$\kappa_{\text{att}}$ (Sec.~\ref{sec:direction}), and budget $K$.
Across Dirichlet prior concentration
$\alpha_0\in\{10^{-3},\ldots,10^{2}\}$, \attn{} maintains
${\sim}0.85$ plateau survival while \base{} collapses at
$\alpha_0\geq 10$.

The budget sweep is more nuanced. Across $K\in\{1.5,\ldots,4.0\}$,
\attn{} beats \base{} by $32$--$56$\,pp at every tested $K$. Both
peak near the canonical $K{=}2.60$ and decline at high $K$.
Selective allocation gives $\kappa_{\text{att}}{=}0.90$ on the
attended channel and
$\kappa_{\text{un}}{=}(K-\kappa_{\text{att}})/3$ on the other
three. At $K{=}4$ the cap binds, so the direction claim is cleanly
identified at the canonical $K{=}2.60$ where no channel saturates
(Sec.~\ref{sec:direction}). The baseline's decline at high $K$
is a planner-overcommitment effect that selective allocation avoids
by keeping the policy posterior softer.

To see what the dynamic selector adds, we compare it against a
fixed-channel control that always $\kappa$-shapes one channel. On
easy-tier layouts,
where hunger dominates the need landscape, always-attend-hunger
ties dynamic \attn{} (both $0.83$). The dynamic selector pulls
ahead in two regimes. Under rigid priors ($\alpha_0\geq 10$),
fixed-channel modes pool to ${\approx}0.65$ while \attn{} reaches
${\approx}0.90$. On a forced-multi-need setting where food- and
water-need alternate in dominance, dynamic beats always-attend-hunger
by pooled $+23.5$\,pp, since no single fixed channel tracks the
switching need. The per-observation learning difference on the
currently-most-needed channel (Sec.~\ref{sec:relevant}) is what
the dynamic selector buys in both regimes.

The advantage also survives environmental non-stationarity: a
tile-mutation sweep confirms it in sign across mutation rates
$0.02$--$0.10$, though it weakens at high rates.

\subsection{Does the Attended Channel Also Learn Faster?}
\label{sec:relevant}

Selective allocation also makes the agent learn the body channel it
attends to faster. Over 50 trials, \attn{}'s hunger model converges
about $2.4\times$ faster than \base{}'s, and this is a per-observation
effect rather than the result of a surviving agent gathering more
data. Plotting hunger-model accuracy against cumulative observation
count rather than trial index, \attn{} sits ${\sim}0.31$ above \base{}
at every matched observation level. The anti-aligned control \rev{}
sits ${\sim}0.22$ above, so any non-uniform allocation buys some
acceleration, but \attn{} still beats \rev{} by a further
${\sim}0.09$, so direction matters per observation as well as per
trial.

This gap is what we would expect if $\kappa$ sharpens the Dirichlet
update at the update step itself, so changing precision shows up not
just in survival but in how fast each channel is learned. We treat
this as an interesting behavioural difference between the two agents
rather than a decisive test of mechanism. A preference-reweighting
scheme that changed only which channel the planner pursues, without
sharpening the per-step likelihood, might behave differently here, and
settling that needs a matched comparator we leave to future work.

\section{Discussion}
\label{sec:discussion}

Selective interoceptive precision is, on its own, enough to
prioritise between competing needs, and the \attn{}'s roughly
twofold survival advantage over the \base{} at matched budget
establishes that. The propagation, direction, and per-observation
learning results then locate where the signal acts and are
consistent with it leaving a behavioural trace beyond survival
(Sec.~\ref{sec:relevant}). The robustness of that advantage comes
from what drives the allocation. Because $\kappa$ is set by the
body-state belief, which updates independently of the world model,
the prioritisation signal survives both rigid-prior world-model
collapse and spatial-map non-stationarity. The body-signal-free
\base{} tolerates neither. The $\kappa$-routing here is
hand-specified, and whether a learned router converges to the same
need-aligned policy is the natural follow-up.

The work draws on two literatures at once, attention-as-precision
in active inference
\cite{parr2017uncertainty,parr2022active,mirza2019introducing} and
active-interoceptive inference
\cite{seth2016active,solms2018hard,pezzulo2015active,tschantz2022simulating,cea2026insentient}.
Within the first, the shaped likelihood matters at the planning
stage and not only at perception, since feeding it into the EFE
planner accounts for about half the benefit at the default prior
and almost all of it under rigid priors. A complementary line
extends active-inference agents at the planning layer through
active learning \selfcitehodson. We make the parallel extension at
the precision-allocation layer.

\begin{sloppypar}
Homeostatic RL~\cite{keramati2014,yoshida2024homeostatic,yoshida2024emergence}
integrates interoceptive state through a hand-designed
homeostatic reward, a connection Keramati and Gutkin
drew explicitly to active inference and interoceptive surprise.
Our contribution routes behaviour through per-step observation
precision rather than through that reward.
\end{sloppypar}

We offer one biological reading tentatively, as a way of thinking
about the model rather than a claim about the brain. It is tempting
to relate the $\kappa$ allocation to the idea that interoceptive
processing is gain-modulated by bodily urgency
\cite{seth2016active,fermin2022insula,livneh2017agrp}. If so,
blunting that gain might impair need prioritisation most while an
agent is still learning an unfamiliar environment, where the
\attn{}--\base{} gap of Table~\ref{tab:headline} is largest. We
intend this as a hypothesis-generating analogy, not a specific
neuroanatomical or clinical prediction.

\section{Limitations and Scope}
\label{sec:limitations}

Our experiments vary the routing with the selector fixed.
The claim is therefore that selective interoceptive precision is
\emph{one sufficient implementation} of the prioritisation
mechanism, not that it is uniquely identified relative to
alternative actuation sites for the same selector. A matched
preference-reweighting comparator is the natural next experiment.
It would test whether the learning difference of
Sec.~\ref{sec:relevant} is driven by per-observation Dirichlet
acceleration rather than by planning-action selection. The selector
also assumes the body-state posterior is well-calibrated to true
need. Under strong miscalibration the same mechanism could misroute
precision, which our fixed-selector design does not test.

The empirical scope is bounded (layout bank, depletion-rate range,
planning horizon, channel count, discrete grid). \textsc{AffectWorld}
is a deliberately minimal gridworld, so the biological readings
above are offered as model-generated hypotheses about mechanism,
not as claims that scale to the full complexity of biological
interoception or to embodied intelligence more broadly.
The \attn{}-vs-\base{} contrasts pass at $\alpha{=}0.01$ with a
comfortable margin.

\section{Conclusion}
\label{sec:conclusion}

\begin{sloppypar}
We asked whether an agent should point a limited perceptual budget
at whichever need is most urgent, and what such routing buys it.
The answer is that selective interoceptive precision, driven by the
agent's own belief about which need is most pressing, is by itself
enough to prioritise between competing needs, roughly doubling
learning-phase survival over a uniform-precision agent at the same
budget. Two features make the result specific rather than generic.
The gain depends on allocating precision at a genuine need, since
reversing the direction does worse than spreading precision evenly.
It works through the planner as much as through perception, because
the shaped likelihood must reach policy evaluation for most of the
benefit to appear.
\end{sloppypar}

\begin{credits}
\subsubsection{Acknowledgements}
\begin{sloppypar}
Funded by the Oppenheimer Memorial Trust (UCT Neuroscience
Institute, 475201 NSI1006) and Conscium, Ltd (UCT Psychology,
PSY526 428430). EAB: NWO grant 019.223SG.002. RS: Laureate
Institute for Brain Research. Computations: UCT ICTS HPC. SG used
Claude (Anthropic) for coding and manuscript editing.
\end{sloppypar}

\subsubsection{Disclosure of Interests}
The authors have no competing interests to declare.
\end{credits}

\bibliographystyle{splncs04unsrt}
\bibliography{paper1_refs}

\clearpage
\appendix
\section*{Supplementary Material}
\addcontentsline{toc}{section}{Supplementary Material}

These supplementary Sections A--E were not part of the 12-page SAB 2026
camera-ready paper. In the combined arXiv document, that paper is reproduced
without alteration and followed by these sections. They report implementation
details and analyses for the same agents, layouts and binary 60-step survival
outcome. We first specify the planner and precision mechanism, then report
layout and statistical sensitivity, selector and propagation controls,
parameter robustness, food-to-poison non-stationarity, and learning at matched
observation count. The agent, environment and layout banks are available at
\url{https://github.com/sgrimbly/attention-aif-sab2026-snapshot}.

\subsection*{A. Implementation Details}

\paragraph{Planner objective.}
The reported experiments used the pragmatic policy score, with state
information gain disabled. The resulting objective was the cross-entropy
\begin{align*}
  \mathcal{G}(\pi) &\;=\; -\sum_{\tau}\sum_{m}
  \mathbb{E}_{q(o^{(m)}_\tau\mid\pi)}\!\left[\log P\!\left(o^{(m)}_\tau \mid C\right)\right] \\[2pt]
  &\;=\; \underbrace{\mathrm{KL}\!\left[q(o\mid\pi)\,\|\,P(o\mid C)\right]}_{\text{risk}}
  \;+\; \underbrace{H\!\left[q(o\mid\pi)\right]}_{\kappa\text{-dependent entropy}} .
\end{align*}
In other words, interoceptive precision still reached the planner through the
predicted-observation distribution
$q(o\mid\pi)=A^{(m)}q(s\mid\pi)$, without a separate
state-information-gain term.

\paragraph{Where precision acts.}
For channel $m$, $\kappa_m$ sets both the probability that the body emits the
true interoceptive level and the precision of the agent's corresponding
likelihood $A^{(m)}$. It therefore changes the sampled observation before
inference, while the same likelihood shapes belief update and policy
evaluation. In this sense, selective precision is a limited sensing capacity
directed towards one channel at a time.

\paragraph{Policy precision.}
The reported runs used the planning library's default policy precision
$\gamma=1$ in $P(\pi)\propto e^{-\gamma\mathcal{G}(\pi)}$. All agents shared
this value, so the reported contrasts are at matched $\gamma$.

Table~\ref{tab:notation} collects the notation used below. The four agent
shorthands are \base{} (uniform $\kappa$), \attn{} (dynamic $\kappa$ towards
the most-needed channel), \rev{} (dynamic $\kappa$ towards the least-needed
channel), and \inferonly{} (the shaped likelihood withheld from the planner).

\begin{table}[!ht]
\centering
\caption{Notation used in the paper and supplement.}
\label{tab:notation}
\begin{tabular}{ll}
\toprule
Symbol & Meaning \\
\midrule
$M$ & number of interoceptive channels ($M{=}4$: 3 active needs + 1 inert control) \\
$m^\star_t$ & attended channel at step $t$ \\
$\kappa_m$ & observation precision on channel $m$ \\
$\kappa_{\mathrm{att}}$ & precision given to the attended channel ($0.90$) \\
$\kappa_{\mathrm{un}}$ & precision on each unattended channel ($0.567$) \\
$K$ & total precision budget, $\sum_m \kappa_m \le K = 2.60$ \\
$A^{(m)}$ & observation likelihood for channel $m$ \\
$B$ & body-state transition prior \\
$C$ & preferred-observation distribution \\
$\mathcal{G}(\pi)$ & policy score \\
$\gamma$ & policy precision ($\gamma{=}1$) \\
$\alpha_0$ & Dirichlet prior concentration for learning $A$ \\
$H$ & planning horizon ($H{=}3$) \\
\bottomrule
\end{tabular}
\end{table}

\subsection*{B. Layout and Statistical Sensitivity}

We group the 12 layouts by the start-to-farther-resource distance. This is
$1$--$2$ cells in the easy tier, $3$--$4$ cells in the medium tier, and $6$
cells in the far tier. \texttt{L01} is not geometrically unique. It is separated
because it is the only easy-tier layout on which the \base{} collapses under a
learned world model while recovering under the true model. This is a post-hoc
behavioural criterion, and excluding \texttt{L01} is conservative because its
inclusion increases the attentive-over-uniform gap.

\begin{figure}[!htb]
\centering
\includegraphics[width=\textwidth]{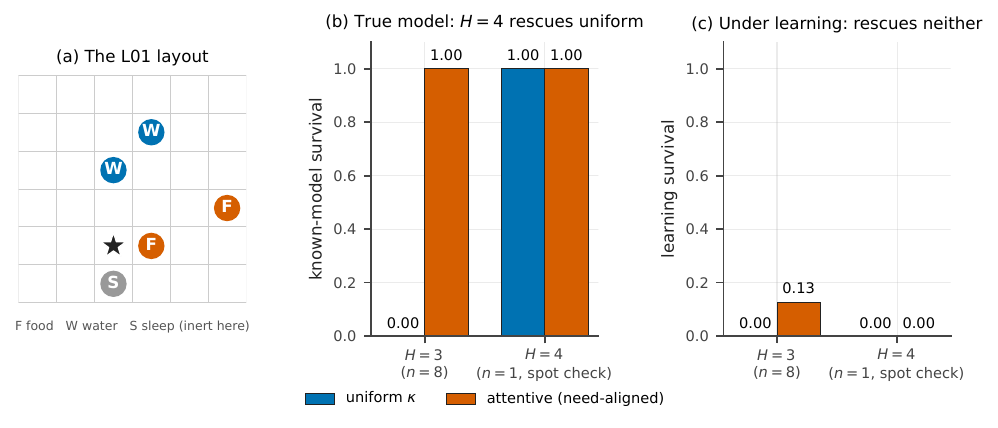}
\caption{\textbf{\texttt{L01} horizon sensitivity.} \textbf{(a)}~The layout and start
position. \textbf{(b)}~With the true world model, increasing the planning
horizon from $H{=}3$ to $H{=}4$ rescues the \base{}. \textbf{(c)}~Under
learning, the same increase does not rescue it, while the \attn{} already
survives at $H{=}3$. The $H{=}3$ cells use $n{=}8$ batched seeds over 50 trials;
the $H{=}4$ cells use one seed over 10 trials and are descriptive only.}
\label{fig:l01}
\end{figure}

On the full 12-layout panel the \attn{} survives at $0.378$ against the
\base{}'s $0.174$, a $+20$ percentage-point difference at
$p\leq10^{-4}$. Since the $H{=}4$ cells use one seed, they do not estimate an
effect size. Instead, they suggest that the \texttt{L01} failure depends on the
start position, shallow planning and a learned world model together.

The headline confidence intervals and tests resample paired (layout, seed)
clusters rather than individual trials, using $10{,}000$ bootstrap resamples.
Holm--Bonferroni correction is applied to the three primary survival contrasts
in Table~\ref{tab:holm}. The remaining sweeps in this supplement are
exploratory.

\begin{table}[!ht]
\centering
\caption{Holm--Bonferroni-corrected paired-bootstrap $p$-values for the three
primary survival contrasts. Values at the bootstrap resolution floor are
reported as $\leq 3\times10^{-4}$.}
\label{tab:holm}
\begin{tabular}{lrrl}
\toprule
comparison (vs \base{}) & $p_\text{raw}$ & $p_\text{adj}$ & reject $H_0$? \\
\midrule
\attn{} survival, easy tier (excl.\ \texttt{L01}) & $\leq10^{-4}$ & $\leq3\times10^{-4}$ & yes \\
\attn{} survival, all 11 layouts (excl.\ \texttt{L01}) & $\leq10^{-4}$ & $\leq3\times10^{-4}$ & yes \\
\rev{} survival, easy tier (excl.\ \texttt{L01}) & $0.0274$ & $0.0274$ & yes \\
\bottomrule
\end{tabular}
\end{table}

\begin{figure}[!htb]
\centering
\includegraphics[width=0.85\textwidth]{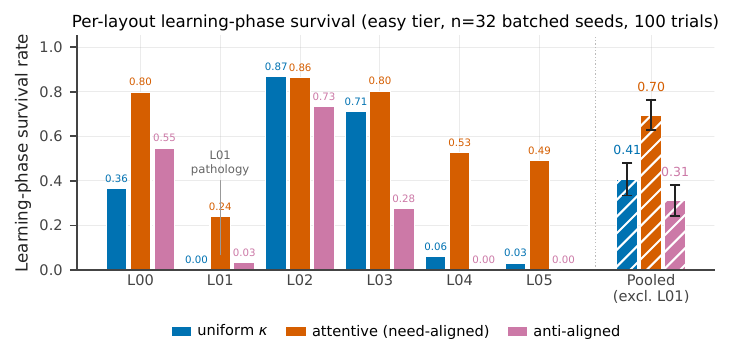}
\caption{\textbf{Easy-tier survival by layout.} Per-layout learning-phase
survival across the six easy-tier layouts, $n{=}32$ batched seeds and 100 trials
per cell. The pooled bars exclude \texttt{L01}. The \attn{} leads the \base{}
on five layouts. On \texttt{L02}, where both are near ceiling, their estimates
differ by one percentage point.}
\label{fig:headline}
\end{figure}

\begin{table}[!ht]
\centering
\caption{Easy-tier sub-panel excluding \texttt{L01}, with five layouts,
$n{=}32$ batched seeds and $16{,}000$ trial outcomes per agent.}
\label{tab:headline-easy}
\begin{tabular}{lccc}
\toprule
Agent & Survival rate & 95\% CI & $p$ vs \base{} \\
\midrule
\Base{} & $0.407$ & $[0.334, 0.482]$ & --- \\
\Attn{} & $\mathbf{0.696}$ & $[0.628, 0.764]$ & $\mathbf{\leq10^{-4}}$ \\
\Rev{} & $0.311$ & $[0.243, 0.383]$ & $0.027$ \\
\bottomrule
\end{tabular}
\end{table}

The ratio of survival rates depends on which layouts are pooled, so the absolute
difference is easier to interpret. The far tier contributes a common zero for
every agent, while the \base{} is also at zero throughout the medium tier. The
attentive-over-uniform difference is $+29$ percentage points on the easy-tier
sub-panel and $+21.5$ percentage points across the 11-layout headline panel.

\subsection*{C. Selector and Propagation Controls}

\begin{figure}[!htb]
\centering
\includegraphics[width=\textwidth]{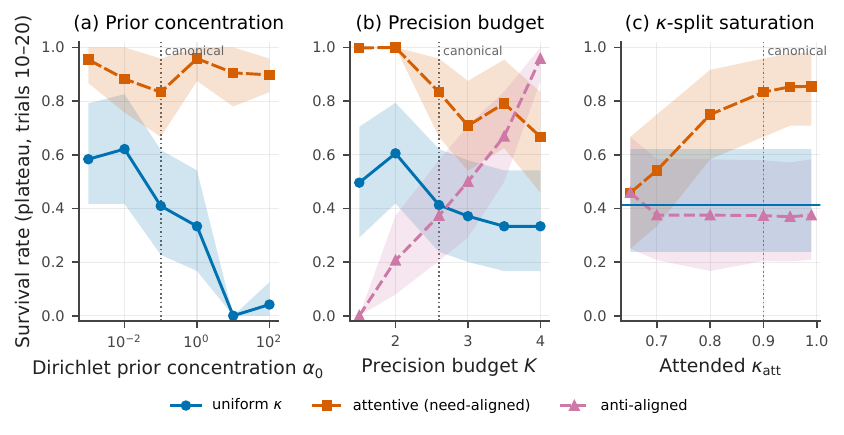}
\caption{\textbf{Robustness to prior, budget, and allocation.} Plateau survival
across three parameter sweeps, with three layouts and $n{=}8$ batched seeds per
cell. \textbf{(a)}~The \attn{} remains stable as Dirichlet prior concentration
increases while the \base{} collapses.
\textbf{(b)}~The \attn{} exceeds the \base{} at every tested precision budget;
the high-$K$ cells enter a saturation regime. \textbf{(c)}~Need-aligned and
anti-aligned selectors coincide at uniform allocation and separate as
$\kappa_{\mathrm{att}}$ increases. Panel \textbf{(a)} uses trials 10--20;
panels \textbf{(b)} and \textbf{(c)} use trials 10--30. Ribbons are
cluster-bootstrap 95\% CIs over (layout, seed).}
\label{fig:robustness}
\end{figure}

The \attn{} and \rev{} use the same precision multiset
$\{0.90,0.567,0.567,0.567\}$ and therefore the same total budget. They differ
only in which channel receives the larger value. The \rev{} performs worse
than the \base{} on the 11-layout panel (raw $p{=}0.004$), so uneven allocation
does not explain the benefit on its own. Panel \textbf{(c)} of
Fig.~\ref{fig:robustness} provides the corresponding control over allocation
magnitude. The selectors coincide at the uniform allocation
$\kappa_{\mathrm{att}}=0.65$ and separate as the allocation becomes sharper.

\begin{figure}[!htb]
\centering
\includegraphics[width=\textwidth]{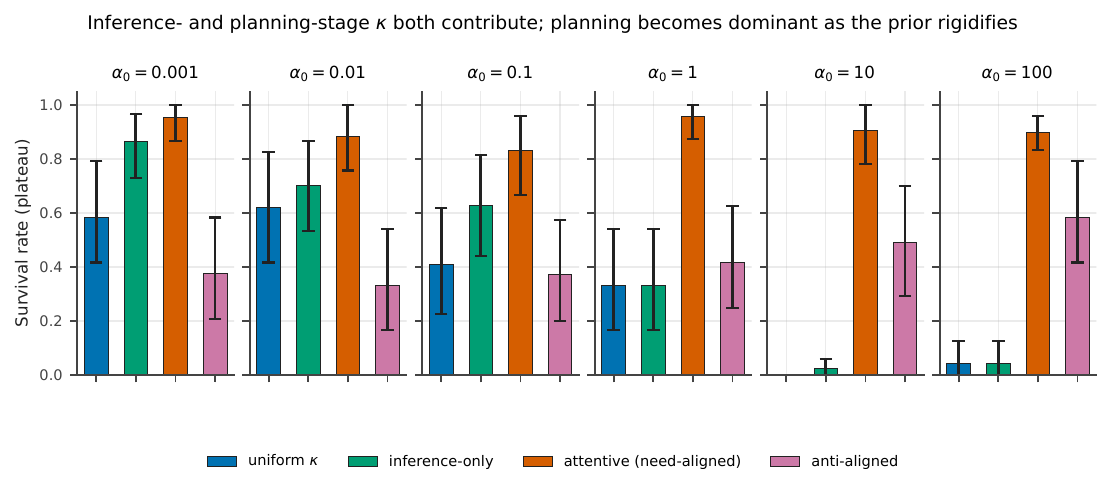}
\caption{\textbf{Where the precision signal acts.} Plateau survival (trials
10--20) across six Dirichlet prior concentrations $\alpha_0$, over three
easy-tier layouts and $n{=}8$ batched seeds per cell. The \inferonly{} withholds
the shaped likelihood from the planner. Its gap from the full \attn{} widens as
the body-state prior becomes more rigid.}
\label{fig:mechanism}
\end{figure}

At $\alpha_0=10$, plateau survival is $0.905$ for the full \attn{} and $0.023$
for the \inferonly{}, close to the \base{} at $0.000$. The \rev{} remains
intermediate at $0.492$. At the default $\alpha_0=0.1$, inference alone retains
more of the benefit. As the prior becomes rigid, the shaped likelihood must
reach the planner for the advantage to remain.

The budget sweep covers $K\in\{1.5,2.0,2.6,3.0,3.5,4.0\}$ and does not plot the
exact allocation-degeneracy point $K=M\kappa_{\mathrm{att}}=3.6$. At that value,
the need-aligned and anti-aligned variants both assign $\kappa=0.90$ to every
channel, so selector direction no longer changes the allocation. The uniform
curve uses the paper's original baseline implementation, so panel \textbf{(b)}
is not a matched-architecture convergence test. At $K=4$, unattended channels
clip at $1.0$, making the nominally attended channel the least precise. These
cells therefore describe saturation rather than a reversal of the mechanism.

\subsection*{D. Food-to-Poison Non-stationarity}

A fixed resource map is the easiest case for a learned spatial model. We relax
this assumption by allowing food to turn into poison in place during a trial.

\begin{figure}[!htb]
\centering
\includegraphics[width=0.85\textwidth]{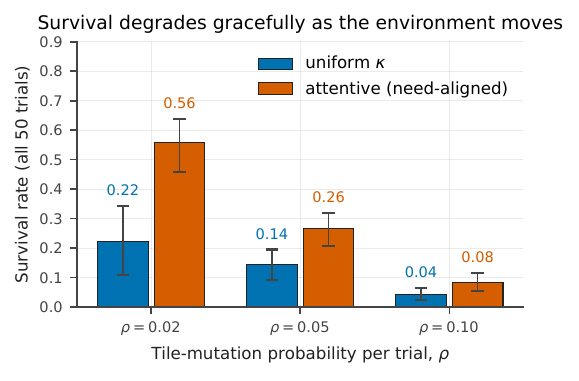}
\caption{\textbf{Food-to-poison non-stationarity.} At each step, a food tile in
the agent's destination cell turns to poison in place with probability $\rho$.
Water tiles never change and no tile moves. The \attn{} leads the \base{} at
all three mutation rates, while survival and the absolute gap decline as
$\rho$ increases. Bars show marginal cluster-bootstrap 95\% CIs over (layout,
seed); Table~\ref{tab:nonstat} reports paired tests.}
\label{fig:nonstat}
\end{figure}

\begin{table}[!ht]
\centering
\caption{Food-to-poison sweep over three layouts, $n{=}8$ batched seeds and 50
trials per cell. CIs are cluster-bootstrap intervals over (layout, seed), and
$p$-values are paired-bootstrap values with Holm--Bonferroni correction across
the three mutation rates.}
\label{tab:nonstat}
\resizebox{\linewidth}{!}{\begin{tabular}{rcccc}
\toprule
mutation rate $\rho$ & \Base{} survival & \Attn{} survival & gap (pp) & adjusted $p$ \\
\midrule
$0.02$ & $0.222$\,[0.112, 0.341] & $\mathbf{0.557}$\,[0.462, 0.639] & $+34$ & $<10^{-3}$ \\
$0.05$ & $0.142$\,[0.092, 0.196] & $\mathbf{0.265}$\,[0.208, 0.318] & $+12$ & $0.002$ \\
$0.10$ & $0.043$\,[0.024, 0.065] & $\mathbf{0.084}$\,[0.054, 0.116] & $+4$ & $0.041$ \\
\bottomrule
\end{tabular}}
\end{table}

The conversion occurs before the tile takes effect, so the agent consumes poison
on the step that triggers the change. The tile then stays poisoned for the rest
of the trial, and layouts reset between trials. Since $\rho$ is conditional on
entering a food cell, agents that forage more often are exposed more often. The
attentive advantage remains positive across the sweep, but decreases as the
resource map becomes less reliable.

\subsection*{E. Learning at Matched Observation Count}

For the channel-wise learning analysis, we use three easy-tier layouts
(\texttt{L00}, \texttt{L02} and \texttt{L04}), $n{=}8$ batched seeds and
$\alpha_0=0.1$.

\begin{center}
\begin{minipage}{\textwidth}
\centering
\includegraphics[width=\linewidth]{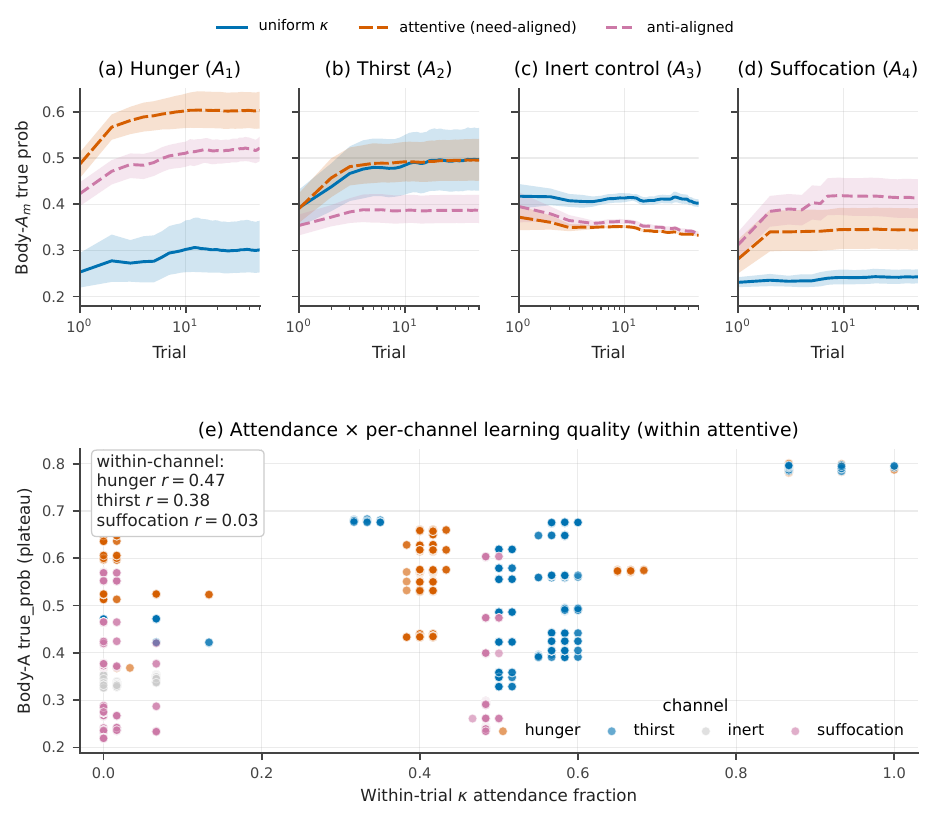}
\captionof{figure}{\textbf{Channel-wise likelihood learning.} Per-channel
Dirichlet learning over 50 trials. \textbf{(a)}~The \attn{}'s hunger model is
more accurate from the first trial ($0.48$ against $0.25$), and both agents
approach their plateaus within about five trials. This is a level difference
rather than a distinct convergence rate. \textbf{(b)}~Thirst learning is similar
under both agents. \textbf{(c)}~The inert control stays near uniform.
\textbf{(d)}~Suffocation shows a smaller difference than hunger.
\textbf{(e)}~Within the \attn{}, attendance fraction correlates with learning
quality for hunger ($r{=}0.47$) and thirst ($r{=}0.38$), but not suffocation
($r{=}0.03$).}
\label{fig:relevant}
\end{minipage}
\end{center}

Over plateau trials, the \attn{} allocates the largest mean share to thirst
($0.49$), followed by hunger ($0.33$), suffocation ($0.17$) and the inert
control ($0.02$). Hunger nevertheless separates the agents most, since it is the
channel the \base{} learns least accurately. The within-channel correlations in
Fig.~\ref{fig:relevant}e are positive for hunger and thirst and near zero for
suffocation, which stays close to its setpoint over most of an episode.

\begin{center}
\begin{minipage}{\textwidth}
\centering
\includegraphics[width=0.85\linewidth]{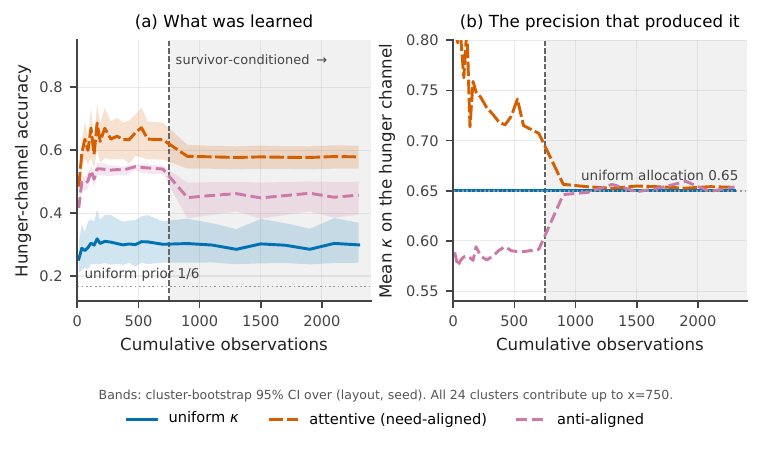}
\captionof{figure}{\textbf{Learning at matched observation count.}
Hunger-channel likelihood accuracy against cumulative observation count rather
than trial index. \textbf{(a)}~Accuracy with cluster-bootstrap 95\% CIs.
\textbf{(b)}~Mean $\kappa$ received by the hunger channel. Where all 24 (layout,
seed) clusters contribute ($x\leq750$), the \attn{}, \base{} and \rev{} receive
mean $\kappa$ of $0.749$, $0.650$ and $0.586$, and reach accuracy of $0.629$,
$0.299$ and $0.524$, respectively. Beyond $x=750$, only longer-surviving
clusters remain and the region is shaded.}
\label{fig:per-obs}
\end{minipage}
\end{center}

The attentive-over-uniform advantage persists at matched observation count, so
it is not a result of longer-surviving agents collecting more data. The \rev{}
remains second in accuracy despite receiving the lowest mean hunger precision.
Direction therefore separates the two dynamic agents more clearly in survival
than in this channel-specific learning measure.

\end{document}